\documentclass[letterpaper, 10 pt, conference]{ieeeconf}  

\usepackage{algorithm2e}
\RestyleAlgo{ruled} 
\usepackage{xcolor}
\usepackage{amsmath}
\usepackage{amssymb}
\usepackage{booktabs}
\usepackage{subcaption}
\usepackage{verbatim}
\usepackage{caption}
\usepackage{hyperref}
\usepackage{cite}

\usepackage{ifthen}
\usepackage{graphicx}
\newboolean{showcomments}
\setboolean{showcomments}{true}

\ifthenelse{\boolean{showcomments}}
  {\newcommand{\nb}[3]{
  {\color{#2}\small\fbox{\bfseries\sffamily\scriptsize#1}}
  {\color{#2}\sffamily\small$\triangleright~$\textit{\small #3}$~\triangleleft$}
  }
  }
  {\newcommand{\nb}[3]{}
  }

\IEEEoverridecommandlockouts                              

\title{\LARGE \bf
Advancing Learnable Multi-Agent Pathfinding Solvers \\ with Active Fine-Tuning
}

\author{Anton Andreychuk$^{1*}$, Konstantin Yakovlev$^{2,1}$, Aleksandr Panov$^{1,3}$ and Alexey Skrynnik$^{1,3*}$
\thanks{*These authors contributed equally to this work}
\thanks{$^{1}$Anton Andreychuk, Konstantin Yakovlev, Aleksandr Panov, and Alexey Skrynnik are with AIRI, Moscow, Russia}%
\thanks{$^{2}$Konstantin Yakovlev is with FRC CSC RAS, Moscow, Russia}%
\thanks{$^{3}$Aleksandr Panov and Alexey Skrynnik are also with Moscow Institute of Physics and Technology, Dolgoprudny, Russia}
\thanks{Work done at the Cognitive AI Systems Laboratory, AIRI.}
}

\begin{document}

\maketitle
\thispagestyle{empty}
\pagestyle{empty}

\begin{abstract}

Multi-agent pathfinding (MAPF) is a common abstraction of multi-robot trajectory planning problems, where multiple homogeneous robots simultaneously move in the shared environment. While solving MAPF optimally has been proven to be NP-hard, scalable, and efficient, solvers are vital for real-world applications like logistics, search-and-rescue, etc. To this end, decentralized suboptimal MAPF solvers that leverage machine learning have come on stage. Building on the success of the recently introduced MAPF-GPT, a pure imitation learning solver, we introduce MAPF-GPT-DDG. This novel approach effectively fine-tunes the pre-trained MAPF model using centralized expert data. Leveraging a novel delta-data generation mechanism, MAPF-GPT-DDG accelerates training while significantly improving performance at test time. Our experiments demonstrate that MAPF-GPT-DDG surpasses all existing learning-based MAPF solvers, including the original MAPF-GPT, regarding solution quality across many testing scenarios. Remarkably, it can work with MAPF instances involving up to 1 million agents in a single environment, setting a new milestone for scalability in MAPF domains.

\end{abstract}

\section{Introduction}

Currently, a wide variety of practical robotic systems involve numerous mobile robots that have to move and operate in the shared environment, such as robots that transport goods in the automated warehouses~\cite{li2021lifelong} or driverless cars in public roads~\cite{li2023intersection}. One of the important abstractions used to formulate and study the problem of finding safe paths in such scenarios is the so-called multi-agent pathfinding (MAPF)~\cite{stern2019multi}. In MAPF, the time is discretized, and agents are confined to a graph (typically a 4-connected grid). Their actions are synchronous, and each action, i.e., moving to an adjacent vertex or staying at the current vertex, takes one time step. The task is to find a set of plans (one for each agent) so that no collisions occur when agents follow them.

Indeed, numerous real-world robotic complications are lifted in MAPF, such as continuous workspace and time, asynchronous actions, communication and observation limitations, perception constraints, etc. Still, MAPF adequately reflects the core challenge of any multi-robot navigation problem: coordinating the agents' actions to avoid collisions and, preferably, optimizing a given cost objective. Consequently, MAPF has gained a lot of attention from the robotics and AI communities. Moreover, numerous works successfully adopt MAPF methods and solutions to the continuous, noisy, and inaccurate world of real robots~\cite{honig2016multi,ma2019lifelong,yakovlev2019prioritized}.

Typically MAPF is solved in a centralized fashion. It is assumed that a single planner exists that fully observes the environment and is responsible for constructing plans for the agents. Both optimal and suboptimal solvers of that kind are known~\cite{standley2010finding,sharon2015conflict,Wagner2011,surynek2016efficient,okumura2022priority,okumura2023lacam,li2022mapf}.
Unsurprisingly, optimal solvers do not scale well to large numbers of agents, as solving MAPF optimally is proved to be NP-Hard~\cite{surynek2010optimization}. Meanwhile, suboptimal solvers scale to hundreds and even thousands of agents while the cost of their solution may degrade significantly, especially for certain setups. Generally, much of the research in the field focuses on striking the right balance between solver's speed and solution quality -- a fundamental trade-off in MAPF. 

One promising approach to achieving this balance is to solve MAPF in a decentralized fashion. Within this approach, MAPF is framed as a decentralized sequential decision-making problem, where at each time step, each agent decides and executes an action based on local observation and communication. The decision-making policy is typically learned or relies on learnable and non-learnable components (hybrid policy)~\cite{liu2020mapper,li2021message,wang2023scrimp,ma2021distributed,ma2021learning,tang2024ensembling,skrynnik2023learn,skrynnik2023switch} (see a recent comprehensive review~\cite{alkazzi2024comprehensive}). 

One of the recent successes in decentralized learnable MAPF is MAPF-GPT~\cite{andreychuk2025mapf}. It relies purely on supervised learning of the transformer-based neural network and utilizes a large dataset of observation-action pairs that consists of approximately 1 billion entries. This way, MAPF-GPT avoids the necessity to carefully hand-craft a reward function as needed in RL-based methods and does not employ a dedicated path-planning module like in hybrid solvers. At the same time MAPF-GPT outperforms numerous learnable MAPF competitors.


Motivated by this, in this work, we explore how the performance of MAPF-GPT can be further improved. To this end, we introduce a novel reward-free fine-tuning method that leverages active learning. In online learning in an environment, we use two types of solvers - approximated to determine the inaccuracies in the current model and accurate to supplement expert data. Our new model, \textbf{MAPF-GPT-DDG}\footnote{The project page:~\href{https://sites.google.com/view/mapf-gpt-ddg}{https://sites.google.com/view/mapf-gpt-ddg}}, is further fine-tuned using new and old expert data. We show that even with a small amount of properly generated training data, fine-tuning a 2-million-parameter (2M) model can achieve the performance of the previously introduced 85-million-parameter (85M) model while requiring significantly lower training costs.

Specifically, we make the following contributions:
\begin{itemize}
    \item We introduce a novel fine-tuning approach called \textbf{Delta Data Generation (DDG)}. The approach significantly speeds up reward-free fine-tuning. 
    
    \item We present \textbf{MAPF-GPT-DDG}, a fine-tuned version of MAPF-GPT, which significantly improves the performance of the original model, establishing a new state-of-the-art for decentralized MAPF solvers. We extensively evaluate its performance, comparing it against other learnable solvers. 
    
    \item \textbf{MAPF-GPT-DDG} can handle MAPF instances involving up to \textbf{1 million agents} in a single environment, setting a new milestone for scalability in MAPF domains.
\end{itemize}

\section{Related Works}

\paragraph{Multi-agent imitation learning (MAIL)} In the realm of multi-agent systems, imitation learning and learning from demonstration are widely employed~\cite{tang2024multiagent,liu2024learning}. 
Within the realm of MAIL, there are various methods to consider, including those that employ Bayesian approaches~\cite{yang2020bayesian}, generative adversarial techniques~\cite{song2018multi,li2024gailpg}, statistical tools for capturing interdependencies between agents~\cite{wang2021multi}, low-rank subspaces~\cite{shih2022conditional}, latent models for coordinating agents~\cite{le2017coordinated}, decision transformers~\cite{meng2023offline}, and more. Demonstrations are frequently used for pretraining in games, such as learnable models for chess~\cite{silver2016mastering, ruoss2024amortized}, and in multi-agent pathfinding tasks, as exemplified by SCRIMP~\cite{wang2023scrimp}. 

\paragraph{Foundation models for multi-agent systems} 
Foundation models are typically trained on large datasets, enabling zero-shot or few-shot learning~\cite{bommasani2021opportunities,yang2023foundation}. For autonomous agents, these models can generalize to new tasks beyond their training, with or without additional demonstrations~\cite{firoozi2023foundation}. Another key feature is their fine-tuning capability, allowing rapid adaptation to specific tasks. While widely used in robotics for multimodal tasks with text-based instructions~\cite{firoozi2023foundation,team2024octo,kim2024openvla}, their application in multi-agent systems remains limited. Notable examples include the Magnetic-One model for language and multimodal tasks in WebArena~\cite{fourney2024magentic} and MAPF-GPT for pathfinding~\cite{andreychuk2025mapf}.

Some papers have explored the possibility of adapting foundation single-agent models for solving multi-agent problems without modifying the pretraining process~\cite{veerapaneni2024work,xu2024multi}. 
This paper investigates fine-tuning foundation models for multi-agent tasks, using MAPF-GPT, to the best of our knowledge, the only available model for action generation in multi-agent environments. No universal pre-trained model exists yet, partly due to the complexity of multi-agent policies and the lack of large expert trajectory datasets for training. The MAPF task serves as a useful testbed for transformer-based foundation models in multi-agent scenarios, offering insights for broader applications.

\paragraph{Fine-tuning of foundation models} As mentioned above, one of the important properties of the foundation pre-trained models is the ability to further fine-tuning. In robotics and for autonomous agents, several approaches have been proposed to fine-tune action models using additional collected demonstration data, either online in a simulator or on a real robot. In one of the early works~\cite{staroverov2023fine}, it was shown that a multimodal transformer model that solves tasks such as visual question answering could be trained to manipulate objects and perform actions on a real robot at the same time. The work~\cite{zhou2024archer} presented an actor-critic architecture with online learning for text tasks. The article~\cite{zhai2025fine} describes a multimodal model based on GT4-V and Gemini, which was trained on additionally collected data. Several papers have focused on transitioning from offline to online learning in robotics applications~\cite{etukuru2024robot}. 
Our work is dedicated to developing a fine-tuning process for multi-agent foundation models and, as far as we know, is the first of its kind.


\paragraph{Multi-agent pathfinding} Several approaches to solving MAPF problems exist. Firstly, there are specialized rule-based MAPF solvers designed to find MAPF solutions quickly, but they do not provide any guarantees regarding the cost of the solutions~\cite{okumura2023lacam,li2022mapf}. Secondly, there are reduction-based approaches that aim to find optimal MAPF solutions by converting the problem into a well-known problem in computer science, such as minimum-flow on graphs or boolean satisfiability (SAT), and then using an existing solver to find the solution to the new problem~\cite{surynek2016efficient}. Thirdly, there are a variety of search-based MAPF solvers that use graph-search techniques to find MAPF solutions~\cite{sharon2015conflict,sharon2013increasing,Wagner2011}. These solvers often provide certain desirable guarantees, such as finding optimal or near-optimal solutions. Additionally, some simple search-based planners lack strong guarantees, such as prioritized planning~\cite{ma2019searching}, which are also widely used. Recently, solvers for the MAPF problem that employ learning techniques have garnered attention. One of the pioneering solvers in this field was PRIMAL~\cite{sartoretti2019primal}, which demonstrated the feasibility of solving the MAPF problem in a decentralized manner using machine learning. The recent learnable MAPF solvers, such as SCRIMP~\cite{wang2023scrimp}, DCC~\cite{ma2021learning}, and Follower~\cite{skrynnik2023learn}, among others, typically utilize reinforcement learning and additional modules, such as communication, to tackle the problem. In contrast to these approaches, our approach relies solely on imitation learning from expert data.

\section{Problem Statement}

MAPF problem is a tuple $(G, v^1_s, ..., v^n_s, v^1_g, ..., v^n_g)$, where $G$ is a graph to which the agents are confined, $\{v^i_s\}$ -- is a set of the start vertices and $\{v^i_g\}$ is a set of the goal vertices. The task is find a set of $n$ plans $Pl=\{pl^i\}$, one for each agent, such that when following them the agents do not encounter conflicts, i.e. do not use the same graph vertex of the same graph edge at the same time step. The cost of the solution is typically measured as either \textit{Sum-Of-Costs}, $SOC(Pl)= \sum_{i=1}^n cost(pl^i)$, or makespan, $msn(Pl)= \max_{i=1}^n cost(pl^i)$, where $cost(pl^i)$ is the time step when agent $i$ reaches its goal and never moves away. In this work we do not aim at solving MAPF optimally but the solutions of the lower costs are preferable.

Alternatively, MAPF can be viewed as a sequential decision making (SDM) problem, when at each time step a centralized policy $\pi_{centralized}$ makes a decision on what (collision-free) joint action $\textbf{a}=a^1\times...\times a^n$ should be performed, where $a^i$ is the action of the $i$th agent. The problem is to construct such a policy (e.g. by hand-crafting the rules, reinforcement learning etc.).

In this work we treat MAPF as a decentralized SDM problem, i.e. we seek to construct an homogeneous individual policy $\pi$ (the same for every agent) that at each timestep decides which action should be performed by an individual agent. By design, we assume (similarly to many other works in the area) that this policy takes as input not the entire state of the environment but rather a local observation of the agent. The latter encompasses the information about the obstacles and the other agents that are located only in some vicinity of the current agent.

\section{Background}

\subsection{Large-Scale Imitation Learning for MAPF}

Imitation learning aims to approximate an expert policy $\hat{\pi}$ by learning a policy $\pi_{\theta}$, parameterized by $\theta$. First, given an expert policy $\hat{\pi}$, a dataset of expert trajectories is collected:

\begin{equation*}
\hat{\mathcal{T}} = \{ \hat{\tau} \}, \quad \hat{\tau} = \{ (s^1, \mathbf{a}^1), (s^2, \mathbf{a}^2), \dots, (s^L, \mathbf{a}^L) \},
\end{equation*}

\noindent where each trajectory $\hat{\tau}$ consists of a sequence of state-action pairs $(s_t, \mathbf{a}^t)$, $\mathbf{a}^t$ is a joint action. In MAPF $\hat{\pi}$ is instantiated as a centralized solver. 

Then individual trajectories are extracted from $\hat{\mathcal{T}}$:
\begin{equation*}
    \tau_u^{\hat{\pi}} = \{ (o^1_u, a^1_u), (o^2_u, a^2_u), \dots, (o^L_u, a^L_u) \},
\end{equation*}

\noindent where $o^t_u$ is the local observation of agent $u$ at timestep $t$, and $a^t_u$ is the corresponding expert action. The observation representation may vary (e.g. it can come as a tensor of numeric values or as a sequence of tokens for GPT-like models~\cite{ruoss2025amortized}. The resulting set of observation-action pairs forms the training set: $\mathcal{D} = \bigcup_u \{ \tau_u^{\hat{\pi}} \}$.

This dataset is further used to learn a policy $\pi_{\theta}$ by minimizing the negative log-likelihood of expert actions under $\pi_{\theta}$. Formally, the optimization objective is:
\begin{equation}
    \theta^\star = \arg\min_{\theta} \mathbb{E}_{(o_u, a_u^{\hat{\pi}}) \sim \mathcal{D}} \left[ -\log \pi_{\theta} (a_u^{\hat{\pi}} \mid o_u) \right].
    \label{eq:cross-entropy}
\end{equation}

After training, actions can be sampled as $a^u \sim \pi_{\theta} (o_u)$.

\subsection{MAPF-GPT}

MAPF-GPT\cite{andreychuk2025mapf} is a recently introduced learnable approach for solving the MAPF problem that relies purely on imitation learning (with no reinforcement learning involved). It uses LaCAM*\cite{okumura2024engineering} as the expert policy to be cloned.

MAPF-GPT is based on a non-autoregressive transformer which is trained on an expert dataset consisting of 1 billion observation-action pairs. These pairs came from running LaCAM* on problem instances on maze-like and random maps. Each observation consists of a local map and agent-specific data. The map includes a local $11\times11$ field of view with cost-to-go values (i.e. the values that encode the path's length from the nearby cells to the goal) and agents' data -- their positions, targets, and action history. The observation is tokenized (i.e. encoded specifically as a sequence of vectors). The vocabulary consists of 67 distinct tokens in total, while the model’s input (context) comprises 256 tokens drawn from this vocabulary. 


Notably, MAPF-GPT, which in a nutshell predicts the next token (action) provided by a sequence of tokens (observation), does not incorporate any ad-hoc centralized collision-resolution techniques, communication blocks, or online search-based guidance. Still, MAPF-GPT demonstrates great performance and scalability surpassing numerous state-of-the-art learning-based MAPF solvers. 


\section{Method}
Our method builds upon MAPF-GPT and is aimed at mitigating one of the principal issues of any imitation learning approach (like MAPF-GPT): the distribution of states encountered by a learned policy $\pi_\theta$ at the test phase may be different from the distribution of states of the expert policy $\hat{\pi}$ encompassed during training. This problem is known as distributional shift.

Specifically, the policy is optimized under the state distribution encountered by the expert policy $\hat{\pi}$, denoted as:
\begin{equation}
P_{\hat{\pi}}(s) = \Pr(s | \hat{\pi})
\end{equation}

However, during execution, the policy $\pi_\theta$ encounters a different state distribution:

\begin{equation}
P_{\pi_\theta}(s) = \Pr(s | \pi_\theta)
\end{equation}

This discrepancy arises due to decentralization, changes in policy input from states to observation, absence of search procedure, and possible generalization errors. If $P_{\pi_\theta}(s)$ deviates significantly from $P_{\hat{\pi}}(s)$, the policy may operate on an unseen distribution of states, leading to poor performance.

\begin{figure*}
    \centering
    \includegraphics[width=1.0\linewidth]{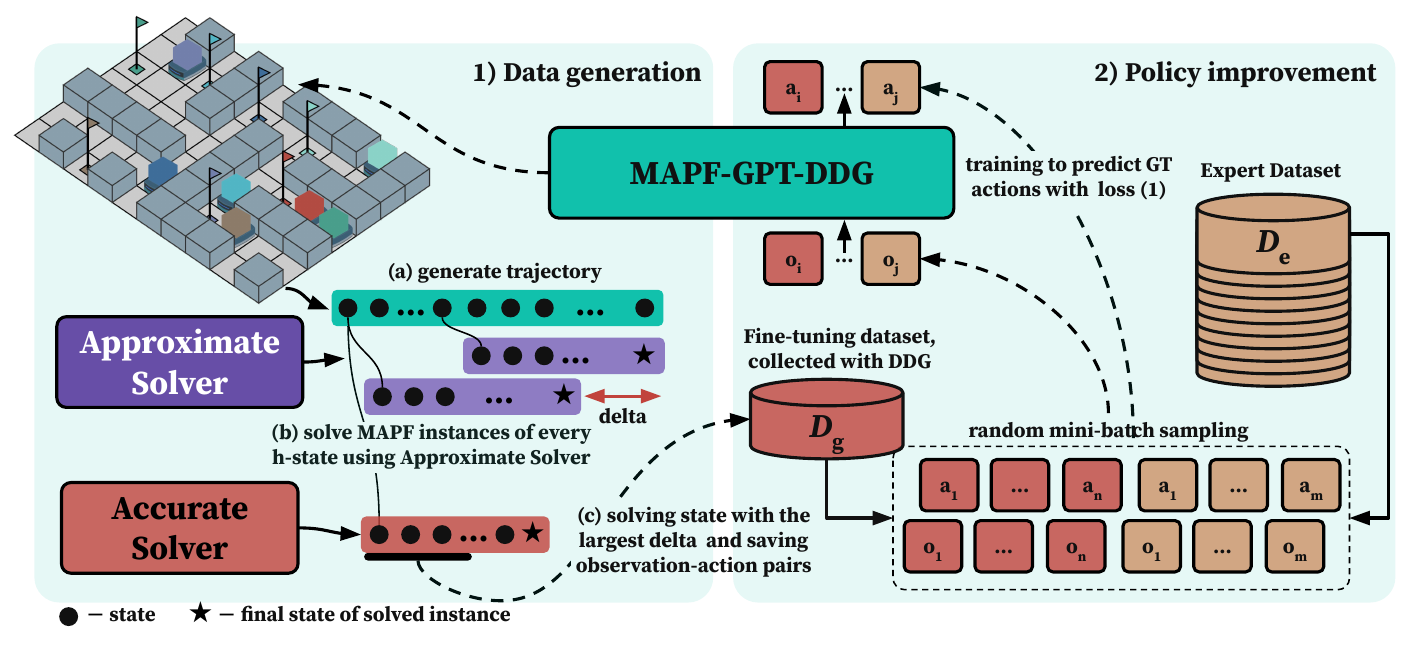}
    \caption{The general pipeline of the proposed active fine-tuning approach — Delta-Driven Data Generation (DDG). The training loop consists of two phases: (1) Data generation, where new trajectories are collected by identifying failure cases of the current policy based on performance degradation (delta) between successive states. The most problematic cases are refined using a more accurate solver, and the corrected examples are added to the training dataset; (2) Policy improvement, where the collected data is used to fine-tune the policy while incorporating the expert dataset to prevent catastrophic forgetting.}
    
    \label{fig:scheme}
\end{figure*}

\subsection{Dataset Aggregation}
A possible way to mitigate the issue of distributional shift is to utilize Dataset Aggregation (DAgger)\cite{ross2011reduction}. DAgger is an iterative imitation learning algorithm designed to address the covariate shift problem, where a learned policy encounters states that deviate from the expert's training distribution. The core idea of DAgger is to iteratively refine the policy by incorporating expert-labeled data on states encountered by the learned policy itself. Specifically, the process works as follows:
\begin{enumerate}
\item Initial Training: The policy $\pi_\theta$ is first trained using standard behavioral cloning on a dataset generated by an expert policy $\hat{\pi}$. This dataset contains state-action pairs $(s, \hat{\pi}(s))$.
\item Execution and Data Collection: The learned policy $\pi_\theta$ is then deployed to interact with the environment. During this process, it encounters a new distribution of states $P_{\pi_\theta}(s)$.
\item Expert Relabeling: To correct for distributional shift, the expert policy $\hat{\pi}$ is queried on these newly encountered states, generating additional expert-labeled data.
\item Dataset Aggregation and Retraining: The newly collected expert-labeled states are aggregated with the existing dataset, and the policy is retrained on this expanded dataset. This process repeats iteratively until convergence.
\end{enumerate}
DAgger effectively helps align the training and execution distributions, reducing compounding errors caused by policy drift. However, despite its  benefits, our experimental evaluation has demonstrated that straightforward application of DAgger to the training process of MAPF-GPT does not yield positive results. To this end we propose a novel technique called Delta Data Generation.

\subsection{Delta Data Generation}
To address the issue of distributional shift we introduce Delta Data Generation (DDG) -- a reward-free fine-tuning method designed to enhance the training process and improve the performance of the trained model. We use this technique to train a new model for MAPF dubbed MAPF-GPT-DDG. The high level scheme of MAPF-GPT-DDG is shown in Fig.~\ref{fig:scheme}, while Algorithm \ref{alg:ddg} provides a formal description.

\begin{algorithm}[ht!]
    \KwIn{$\pi_\theta$, $\mathcal{D}_e$, $\mathcal{G}$, $L$, $J$, $\delta_{min}$, $h$, $k$, $size$, $Solver_{approximate}$, $Solver_{accurate}$}
    $\mathcal{D}_g \gets \emptyset$\\
    \For{$l = 1$ to $L$}
    {
        \textcolor{blue}{\texttt{// Generate dataset for fine-tuning}} \\
        \While{$|\mathcal{D}_g| < size*l$}{
            $\langle s,g \rangle \gets$ generate new instance by $\mathcal{G}$\\
            $S_{\pi_\theta} \gets$ run $\pi_\theta$ on $\langle s,g\rangle$ \\
            $\{s_1, ..., s_n\} \gets$ Extract every $h$-th state of $S_{\pi_\theta}$\\
            \For{$i=1..n-1$}{
                $sol_i \gets$ Solve $\langle s_i,g\rangle$ via $Solver_{approximate}$\\
                $\delta_i \gets cost(sol_{i+1}) - cost(sol_i)$\\
            }
            $z \gets arg\max_{i=1..n} \delta_i$\\
            
            \If{$\delta_z > \delta_{min}$}
            {
                $sol_z \gets$ Solve $\langle s_z,g\rangle$ via $Solver_{accurate}$\\
                $data \gets$ generate observation-action pairs from first $k$ states $s\in sol_z$\\
                $\mathcal{D}_g \gets$  $\mathcal{D}_g \cup data$\\
            }
        }
        \textcolor{purple}{\texttt{// Improve $\pi_\theta$}} \\
        \For{$j = 1$ to $J$}{
            Sample mini-batch $\mathcal{B}_e \sim \mathcal{D}_e$. \\
            Sample mini-batch $\mathcal{B}_g \sim \mathcal{D}_g$. \\
            Combine mini-batches: $\mathcal{B} \gets \mathcal{B}_e \cup \mathcal{B}_g$. \\
            Update policy $\pi_\theta$ using loss eq. (\ref{eq:cross-entropy}) on $\mathcal{B}$.
        }
    }
    \Return Fine-tuned policy $\pi_\theta$
    \caption{Delta Data Generation}
    \label{alg:ddg}
\end{algorithm}

MAPF-GPT-DDG is comprised of two major stages: generating new data and fine-tuning the policy. The data generation process starts by creating a new instance with an initial state $s$ and a goal state $g$ using an instance generator $\mathcal{G}$. Next, a sequence of states $S_{\pi_\theta}$ is generated using the policy $\pi_\theta$. A subset of candidate states ${s_1, ..., s_n}$ is then extracted by selecting every $h$-th state from $S_{\pi_\theta}$. For each extracted state $s_i$, a fast, approximate MAPF solver, $Solver_{approximate}$, computes a solution $sol_i$, and its associated cost, $cost(sol_i)$, is recorded. The cost difference between successive solutions is then computed as:

\begin{equation} \delta_i = cost(sol_{i+1}) - cost(sol_i) \end{equation}

The state $s_z$ that maximizes $\delta_i$ is identified, indicating a scenario where the policy $\pi_\theta$ underperforms. If $\delta_z$ exceeds a predefined threshold $\delta_{min}$, a high-quality solution $sol_z$ for the MAPF instance starting at $s_z$ is computed using an accurate solver, $Solver_{accurate}$. Finally, observation-action pairs generated from the first $k$ states of the improved solution are added to the generated dataset $\mathcal{D}_g$. 


Once the dataset $\mathcal{D}_g$ gets the required amount of data utilizing the current policy $\pi_\theta$, the model undergoes fine-tuning through iterative training. Importantly, while fine-tuning, the expert dataset $\mathcal{D}_e$ is used as well. This is done to prevent the fine-tuned policy to deviate largely from the initial one, the problem known in machine learning as catastrophic forgetting. 

Specifically, during each fine-tuning iteration, mini-batches $\mathcal{B}_e$ and $\mathcal{B}_g$ are sampled from the expert dataset $\mathcal{D}_e$ and the generated dataset $\mathcal{D}_g$, respectively. The combined mini-batch:
$\mathcal{B} = \mathcal{B}_e \cup \mathcal{B}_g$
is used to update the policy $\pi_\theta$ by minimizing loss (eq.\ref{eq:cross-entropy}). This process iterates over $I$ fine-tuning steps per data generation cycle, progressively improving the policy by incorporating corrective feedback from $\mathcal{D}_g$. After that, the loop repeats, until the number of generation iteration $G$, or reaching the time budget.


\subsection{DAgger vs DDG}
Both DAgger and DDG mitigate the distributional shift by iteratively querying an expert for corrections on newly encountered states. Like DAgger, DDG is an online approach that refines the training data over multiple iterations. However, instead of collecting expert data for all encountered states, DDG selectively focuses on the most critical states -- those where the policy demonstrates poor performance. Moreover, DAgger relies on direct expert supervision at every iteration, which can be inefficient, especially in MAPF settings, where calling an expert that is able to provide high-quality solution is resource demanding. In contrast, DDG identifies hard cases by running fast approximate solver, collecting corrective data only where the learned policy struggles. As a result, DDG reduces redundant expert queries while still improving the policy.

\subsection{Training and Parameters Details}
To apply DDG in practice, several choices must be made. First, it is necessary to select appropriate MAPF solvers. Following the original MAPF-GPT in this work, we used LaCAM*\cite{okumura2024engineering}. This is an anytime algorithm that is able to obtain solutions quickly and, when given more time, improve the initial solution. To get fast approximate MAPF solutions, we invoked LaCAM* with a time limit of 2 seconds, and we set this limit to 10 seconds when needed to get higher quality solutions.

The value of $h$ (split length) was set to $16$, $k$ (store length) to $32$, and $\delta_{min}$ (delta threshold) to $3$. These parameters were adjusted to balance the time required for data collection and training. During a single data collection phase, we collected 409,600 observation-action pairs, and one phase of fine-tuning run for 1,000 iterations. The ratio between $\mathcal{B}_g$ and $\mathcal{B}_e$ was set to 1:3, meaning 25\% of data in every mini-batch comes from dataset $\mathcal{D}_g$. 
During the initial iterations, until $\mathcal{D}_g$ reaches sufficient size, we sample fewer data points from it to ensure each observation-action pair is sampled only once during a single fine-tuning phase, avoiding overfitting. As the dataset $\mathcal{D}_g$ grows, we discard older data and retain only the samples generated during the most recent iterations. $\mathcal{D}_g$ was collected on the same type of maps as $\mathcal{D}_e$ -- random and maze maps with a ratio of 1:9 in the resulting data. The maps are generated randomly and guaranteed to differ from those used for experimental evaluation. The number of agents in the instances was set to 32.

To train MAPF-GPT-DDG we chose to fine-tune the 2M model of MAPF-GPT, whose weights are publicly available on Huggingface\footnote{https://huggingface.co/aandreychuk/MAPF-GPT}. It was trained for an additional 340,000 iterations, taking 77 hours on a server with 4 NVIDIA H100 GPUs and 48 CPU cores. This time could be reduced by almost half when running dataset collection and fine-tuning phases asynchronously, as the former primarily relies on CPU resources to run expert while the latter requires mainly GPU computation.

\section{Experimental Evaluation}

\begin{figure*}[ht!]
    \centering
    \includegraphics[width=\textwidth]{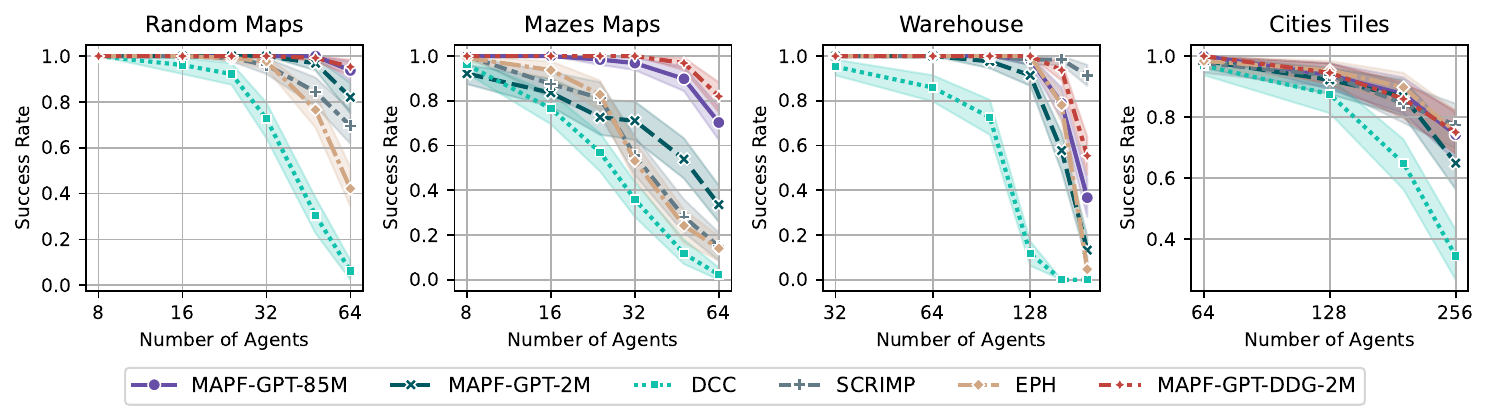}
    \caption{Success rates of the approaches on different map types depending on the number of agents in the instances (higher is better). The shaded area indicates the 95\% confidence interval.}
    \label{fig:csr_plot}
\end{figure*}

\begin{figure*}[ht!]
    \centering
    \includegraphics[width=\textwidth]{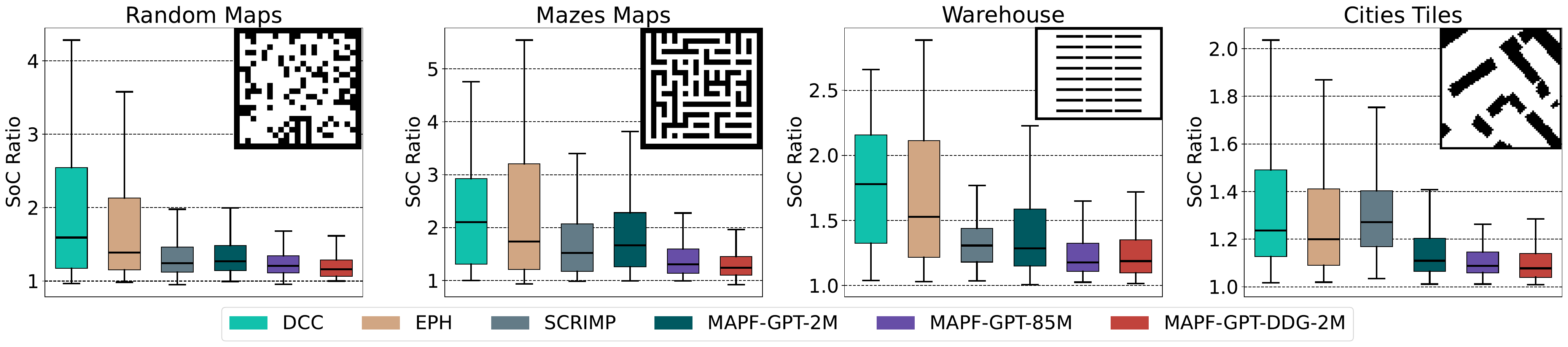}
    \caption{SoC ratio relative to solutions found by the LaCAM* approach (lower is better).}
    \label{fig:soc_plot}
\end{figure*}

\subsection{Experimental Evaluation Setup}
To evaluate MAPF-GPT-DDG empirically, we conducted multiple series of experiments. The main series aimed at face-to-face comparison of MAPF-GPT-DDG to state of the art in learnable MAPF, specifically to the original MAPF-GPT, SCRIMP, DCC, and EPH. The notable original MAPF-GPT comes in three different sizes: 2M, 6M, and 85M, we used 2M (the same size as ours) and 85M models. The experiments were conducted on the POGEMA benchmark~\cite{skrynnik2025pogema} with the same evaluation protocol as in the original MAPF-GPT paper. Specifically, we used 4 map types: Random, Mazes, Warehouse, and Cities Tiles. The first two are the same type of maps that were used to train MAPF-GPT and MAPF-GPT-DDG, the latter two differ significantly in topology and are use to evaluate the ability to generalize to out-of-distribution map types. Mazes and Random maps range in size from $17\times17$ to $21\times21$, and contain up to 64 agents. The Warehouse type features a single map of size $33\times46$ with restrictions on where start and goal locations can be placed (to model real-world fulfillment scenarios). The maximum number of agents on this map is 192. The Cities Tiles maps are of $64\times64$ size, allowing for up to 256 agents. In all runs the episode length was limited to 128 steps, except for Cities Tiles, where the episode length was 256. More details about the benchmark and evaluation protocol can be found in~\cite{skrynnik2025pogema}.

We also performed two additional series of experiments. One studied the influence of the proposed DDG approach on the training process. The other demonstrated the great scalability of MAPF-GPT-DDG and its capability to handle instances with more than a million agents.

\subsection{Comparison with the Baselines}
The first series of experimental results is shown in Fig.~\ref{fig:csr_plot} and Fig.~\ref{fig:soc_plot}. In the former, the average success rate for each map type is presented based on the number of agents in the instances. As can be seen, on Mazes and Warehouse maps, MAPF-GPT-DDG significantly outperforms MAPF-GPT-85M, while on Random and Cities Tiles maps its results are on par with MAPF-GPT-85M -- a model with 50 times more parameters that required 3 times more GPU-hours to train. Comparing with other competitors, SCRIMP demonstrates the best performance, achieving the best results on Warehouse maps. However, on Mazes and Random maps its results are significantly worse. On Cities Tiles maps, all approaches except DCC demonstrate quite close results.

Fig.~\ref{fig:soc_plot} presents the ratio of SoC (solution cost) relative to the solution found by the centralized planner, LaCAM*, in the form of box-and-whiskers plots. These results align with those presented in Fig.~\ref{fig:csr_plot} and show that MAPF-GPT-DDG achieves results close to or slightly better than MAPF-GPT-85M while outperforming all other competitors. Among non-MAPF-GPT solvers SCRIMP demonstrates the best performance. However, it struggles with high-density agent scenarios (such as instances with 256 agents on cities maps).

\subsection{DDG Ablation Study}

In the next series of experiments, we aimed to demonstrate the advantage of utilizing the DDG approach for training MAPF-GPT. For this purpose, we selected the original weights of MAPF-GPT-2M and continued training it in three ways: standard training (using only the precollected 1B dataset), DAgger, and DDG. During this ablation study, we ran DAgger in the same manner as DDG, collecting 409,600 observation-action pairs per single dataset collection phase. In contrast to DDG, the logic of DAgger assumes that expert actions for only the current step are utilized (i.e., its store length is 1). Thus, when launching the expert solver, DAgger collects only 32 (number of agents) observation-action pairs per launch. To expedite this process and make it comparable to the time spent by DDG, the runtime of the expert solver was limited to 2 seconds (the same as the fast solver in the DDG scheme).

\begin{figure}[ht!]
    \centering
    \begin{subfigure}[b]{0.46\linewidth}
        \centering
        \includegraphics[width=\linewidth]{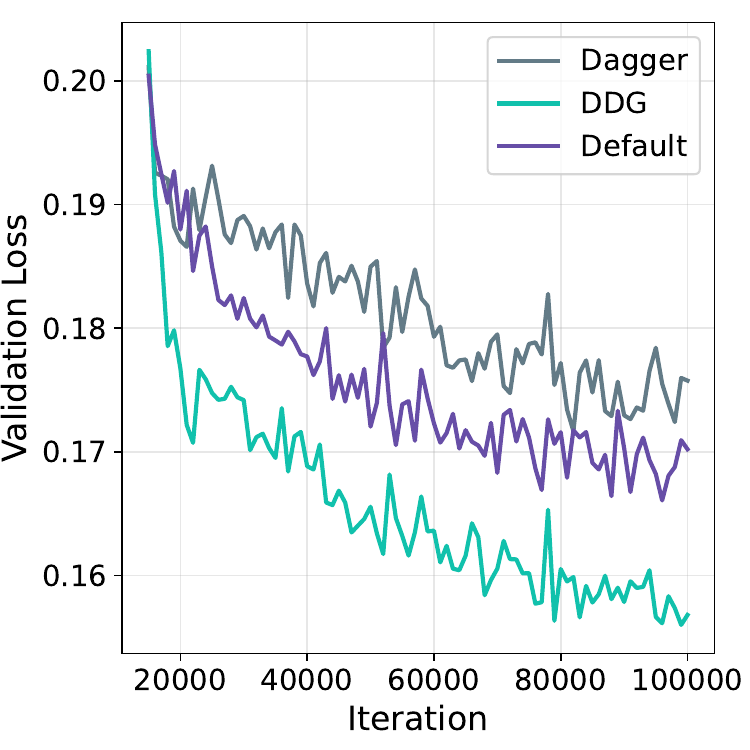}
    \end{subfigure}
    \hspace{8px}
    \begin{subfigure}[b]{0.46\linewidth}
        \centering
        \includegraphics[width=\linewidth]{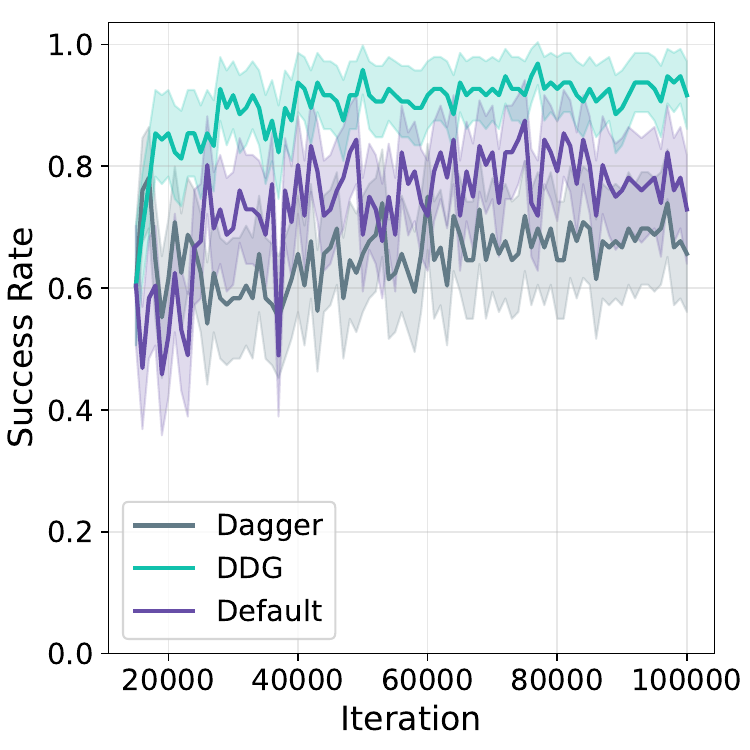}
    \end{subfigure}
    \caption{Validation loss and success rate on subset of instances during the training process.}
    \label{fig:ablation_training}
\end{figure}

Fig.~\ref{fig:ablation_training} illustrates the dynamics of the loss value on the validation portion of the expert dataset, as well as the success rate dynamics on a subset of instances on maze maps with 32, 48, and 64 agents. Both plots indicate that the DDG approach outperforms in both criteria: its loss decreases much more rapidly, and its success rate reaches levels unattainable by other training techniques. Surprisingly, DAgger performs even worse than the default training. We hypothesize that this behavior is due to the fact that many complex situations, where agents become stuck, require a sequence of correct actions to resolve, while DAgger only records the first action. Consequently, the dataset $\mathcal{D}_g$ is even less effective than dataset $\mathcal{D}_e$ for training agents to resolve complex situations. 

\subsection{Demonstration of Scalability}
To demonstrate the exceptional scalability of MAPF-GPT-DDG, we ran it on a large empty map sized $2048\times2048$ with up to $2^{20}$ agents. 
Start and goal locations were generated randomly with a restriction limiting the distance between them to 64. This restriction was implemented to make the instances completable within a viable number of steps. Otherwise, the time required to run such an experiment would increase by 10-100 times.

Table~\ref{tab:million} shows that MAPF-GPT-DDG is able to solve an instance with more than half a million agents and scales linearly with the number of agents. 
The independent success rate indicates that in the instance with $2^{20}$ agents most of them have reached their goals. However, it is not 100\%, that means that there were few agents that got stuck, as a result the task was not solved completely.
Decision time -- the time required for each agent to determine its next action -- is approximately 160 microseconds regardless of the number of agents. To the best of our knowledge, this is the first time instance with such large number of agents is successfully handled by a learnable solver. It is also worth noting that one of the reasons of MAPF-GPT-DDG success in this experiment is that some of its core routines (i.e. tokenization) were reimplemented in C++.

\begin{table}[htb!]
    \centering
    \begin{tabular}{rcrrc}
        \toprule
        Number& Independent  & Episode & {Total} & {Decision} \\
        of Agents& Success Rate $\uparrow$ & Length $\downarrow$ & {Time (s)} $\downarrow$ & {Time ($\mu$s)} $\downarrow$ \\
        \midrule
        8,192   & 100\% & 89  & 117.4   & 161.0 \\
        16,384  & 100\% & 86  & 229.3   & 162.8 \\
        32,768  & 100\% & 100 & 533.2   & 162.7 \\
        65,536  & 100\% & 99  & 1,057.6 & 163.0 \\
        131,072 & 100\% & 92  & 1,941.2 & 163.4 \\
        262,144 & 100\% & 92  & 3,935.7 & 163.2 \\
        524,288 & 100\% & 128 & 10,968.8 & 163.4 \\
        1,048,576 & 99.9\% & 512 & 87,778.4 & 163.5 \\
        \bottomrule
    \end{tabular}
    \caption{Results of MAPF-GPT-DDG on $2048\times2048$ map with up to $2^{20}$ agents.}
    \label{tab:million}
\end{table}

\section{Conclusion}

In this work, we examined how state-of-the-art learnable MAPF solver based on imitation learning, i.e. MAPF-GPT can be improved to achieve an unprecedented performance. Specifically, we have proposed a new approach for active fine-tuning called Delta Data Generation that is tailored to cope with the problem of distribution shift. Unlike previous approaches (like DAgger) DDG focuses on the critically important states where the learnable policy underperforms. We have shown that the solver that was trained using DDG -- MAPF-GPT-DDG -- outperforms the competitors across a large variety of problem instances both in terms of success rate and solution cost. In fact, MAPF-GPT-DDG, which is a 2M model, is able to outperform in certain scenarios an original variant of MAPF-GPT that contains 85M parameters. MAPF-GPT-DDG is a highly efficient solver in terms of computations. We demonstrate it by running MAPF-GPT-DDG on a large map of $2048\times2048$ with up to $2^{20}$ agents. To the best of our knowledge, no one has yet evaluated a learnable MAPF solver on an instance with such a high number of agents.

\bibliography{bib}
\bibliographystyle{IEEEtran}





\end{document}